# Convolutional Factor Graphs as Probabilistic Models

**Yongyi Mao**
School of Information
Technology and Engineering
University of Ottawa
yymao@site.uottawa.ca

**Frank R. Kschischang**
Department of Electrical
and Computer Engineering
University of Toronto
frank@comm.utoronto.ca

**Brendan J. Frey**
Department of Electrical
and Computer Engineering
University of Toronto
frey@psi.toronto.edu

**Abstract**

Based on a recent development in the area of error control coding, we introduce the notion of *convolutional factor graphs* (CFGs) as a new class of probabilistic graphical models. In this context, the conventional factor graphs are referred to as multiplicative factor graphs (MFGs). This paper shows that CFGs are natural models for probability functions when summation of independent latent random variables is involved. In particular, CFGs capture a large class of linear models, where the linearity is in the sense that the observed variables are obtained as a linear transformation of the latent variables taking arbitrary distributions. We use Gaussian models and independent factor models as examples to demonstrate the use of CFGs.

The requirement of a linear transformation between latent variables (with certain independence restriction) and the observed variables, to an extent, limits the modelling flexibility of CFGs. This structural restriction however provides a powerful analytic tool to the framework of CFGs; that is, upon taking the Fourier transform of the function represented by the CFG, the resulting function is represented by a MFG with identical structure. This Fourier transform duality allows inference problems on a CFG to be solved on the corresponding dual MFG.

## 1 INTRODUCTION

Probabilistic graphical models are an effective and efficient methodology for the representation and interpretation of statistical relationship and for statistical inference and learning. Classically, popular choices of such models are *Bayesian networks* (BNs) [1] and *Markov random fields* (MRFs) [2]. For a set of random variables $X_V := \{X_i : i \in V\}$ indexed by a finite set $V$, a BN uses a directed graph with vertex set $V$ to represent the joint probability function [1] $p_{X_V}(x_V)$ of these random variables, in which $p_{X_V}(x_V)$, with respect to the graph, factors as

$$p_{X_V}(x_V) = \prod_{u \in V} p_{X_u|X_{\pi(u)}}(x_u|x_{\pi(u)}).$$

Here $\pi(u)$ is the set of parents of vertex $u$ on the graph, and for any subset $S \subseteq V$, we have used $X_S$ to denote the set of random variables $\{X_i : i \in S\}$ and the corresponding lower-cased symbol $x_S$ to denote the vector-valued configuration that $X_S$ may take. An MRF on the other hand uses an undirected graph representation for the joint probability $p_{X_V}(x_V)$. That is, an MRF represents $p_{X_V}(x_V)$ as

$$p_{X_V}(x_V) = \frac{1}{Z} \prod_{C \in Q} \phi_C(x_C),$$

where $Q$ is the set of maximal cliques of the graph, $\phi_C(x_C)$ is a positive function (known as the potential function) on clique $C$, and $1/Z$ is a scaling factor.

Recently the notion of *factor graphs* [3], referred to as *multiplicative factor graphs* (MFGs) in this paper, has been introduced as a probability model. Given a factorization of the joint probability function $p_{X_V}(x_V)$ into arbitrary functions, an MFG (see its definition in Section 2) uses a bipartite graph to explicitly represent not only the variables but also the factors. More recently, the framework of MFGs is further extended to allow edges to be directed [4]. It is shown that this extended notion of MFG unifies BNs and MRFs and have advantages over both representations.

In this paper, staying with the undirected factor graph formalism, we introduce *convolutional factor graphs*

---

[1] In this paper, we use the term "probability function" in place of "probability mass function" and "probability density function".

(CFGs) as a new class of probabilistic graphical models. The foundation of this paper is presented in [5] and [6], where a generalized notion of multi-variate convolution and the notion of CFGs were presented in the context of error correction coding.

In this paper, we focus on the aspect of using CFGs as probability models. We show that CFGs become natural representations of probability distributions when summation of independent latent random variables is involved. In particular we show that summation of independent latent variables is the necessary and sufficient condition for the convolutional factorization of the probability function of the observed variables.

As an example, we show that every multi-variate Gaussian density factors convolutionally and therefore can be represented by a CFG. More interestingly, in this case, the structure of the CFG precisely corresponds to the covariance matrix of the Gaussian density. In fact, beyond Gaussian models, a large class of probability models may have corresponding CFG representations. To demonstrate this, we derive the CFG representation for the independent factor (IF) model.

In the context of probability models, the global Markov property is known to hold on a MFG and on a (positively constrained) MRF [7]. That is, for every disjoint subsets $X_A, X_B, X_S$ of the random variables on the graph, if $X_S$ separates $X_A$ from $X_B$, then $X_A$ is independent of $X_B$ conditioned on $X_S$. On a CFG, we show that a somewhat dual statement holds. That is, for every disjoint subsets $X_A, X_B, X_S$ of the random variables on the graph, if $X_S$ separates $X_A$ from $X_B$, then $X_A$ is independent of $X_B$ marginally.

Similar to BNs and MRFs, CFGs have its limitation in its modelling flexibility. More specifically, this limitation is that the observed variables must be obtained as a linear transformation of latent variables. In this respect, CFGs are more restricted than BNs and MRFs. This restriction however provides an extra structure to the model so that powerful tools of the Fourier analysis are applicable. That is, by taking the Fourier transform of the function represented by a CFG, the resulting function is readily represented by a MFG with identical structure. As an example, this Fourier transform duality allows inference problems associated with a CFG model to be solved on the corresponding MFG via the Fast Fourier Transform (FFT). This provides a reduced computational complexity comparing with solving the problem without identifying the convolutional factorization of the probability function.

In Section 2, we review the generalized notion of convolutional factorization and the notions of CFGs and MFGs, according to [6]. In Section 3, we develop the general setting for which CFG models are naturally applicable. In Section 4, we discuss inference algorithms based on CFG models.

## 2 FACTORIZATION AND FACTOR GRAPHS

In this section, we summarize the results in [6] that are necessary for this paper. Further details concerning this section may be found in [6].

### 2.1 Convolutional Factorization

Let $f(x_1, x_2)$ and $g(x_2, x_3)$ be two functions. The convolution $f * g$ of $f$ and $g$ is a function involving the union of their variables, defined as

$$(f * g)(x_1, x_2, x_3) := \sum_{y_2} f(x_1, x_2 - y_2) g(y_2, x_3).$$

Here we have assumed that the domain of the functions is discrete; this definition can obviously be extended to functions defined on continuous domains, in which summation is replaced with integration[2]. We remark that this notion of convolution in fact generalizes the conventional multi-variate convolution known in the literature of multi-dimensional signal processing. One may also interpret this notion of convolution as the conventional convolution with respect to the common variables between the involved functions, while treating other variables as the parameters of the functions. We will also write $(f * g)(x_1, x_2, x_3)$ as $f(x_1, x_2) * g(x_2, x_3)$ to remind the readers of the variables of the original functions. Furthermore, if two functions do not have variables in common, their convolution reduces to multiplication.

It can be shown that convolution according to this definition is commutative and associative. This allows us to write the convolution of arbitrary number of functions in any order. In particular, we may use $\prod_{j=1...m}^{*} f_j(x_{S_j})$ to denote convolutional factorization

$$f_1(x_{S_1}) * f_2(x_{S_2}) * \ldots * f_m(x_{S_m}),$$

where $x_{S_j}$ is the set of variables involved in function $f_j$.

### 2.2 MFGs and CFGs

Now we define the two types of factor graphs.

---

[2]More rigorously speaking, as long as each variable takes value from a locally compact abelian group, including the fields of real and complex numbers, integers, finite abelian groups, Euclidean n-space etc, one can define such a notion of convolution.

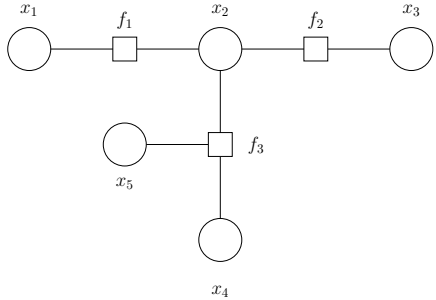

Figure 1: The bipartite graph in Example 1.

A factor graph $(\mathcal{G}, \circ)$ is a bipartite graph $\mathcal{G}$ together with a product operation $\circ$, which is either defined as multiplication $\times$ or convolution $*$; the bipartite graph $\mathcal{G}$ consists of two types of vertices, the set of function vertices, representing a set of functions $\{f_j : j \in J\}$ and the set of variable vertices, representing a set of variables $x_V := \{x_i : i \in V\}$; each function $f_j$ takes a subset $x_{S_j}$ of variable set $x_V$ as its argument, and the union $\bigcup_{j \in J} x_{S_j}$ of these subsets is $x_V$; if $x_i \in x_{S_j}$, then there is an edge connecting variable vertex $x_i$ to function vertex $f_j$. The factor graph $(\mathcal{G}, \circ)$ then represents the product $\prod_{j \in J}^{\circ} f_j(x_{S_j})$. If $\circ$ is defined as convolution $*$, the product is the convolutional product of these functions and the factor graph $(\mathcal{G}, *)$ is said to be a convolutional factor graph (CFG); if $\circ$ is defined as multiplication $\times$, the product is the ordinary (multiplicative) product [3] of these functions and the factor graph $(\mathcal{G}, \times)$ is said to be a multiplicative factor graph (MFG). An example should make this definition clear.

**Example 1** *The bipartite graph $\mathcal{G}$ in Figure 1 consists of variable vertices $x_1, x_2, x_3, x_4$ and function vertices $f_1(x_1, x_2), f_2(x_2, x_3)$ and $f_3(x_2, x_4, x_5)$). Then associating with $\mathcal{G}$ the multiplicative product operation $\times$ defines multiplicative factor graph $(\mathcal{G}, \times)$, which represents function $f_1(x_1, x_2) f_2(x_2, x_3) f_3(x_2, x_4, x_5)$; likewise, associating with $\mathcal{G}$ the convolutional product operation $*$ defines convolutional factor graph $(\mathcal{G}, *)$, which represents function $f_1(x_1, x_2) * f_2(x_2, x_3) * f_3(x_2, x_4, x_5)$.*

### 2.3 Duality

With the classic notion of convolution, it is well-known that the Fourier transform of a convolutional product is a multiplicative product and *vice versa*. This result, known as the *convolution theorem*, in fact holds for this generalized notion of convolution. More precisely, for a set of functions $\{f_j(x_{S_j}) : j \in J\}$, let the Fourier trans-form of $f_j(x_{S_j})$ be $\hat{f}_j(\hat{x}_{S_j})$ for each $j$, then $\prod_{j \in J}^{*} f_j(x_{S_j})$ and $\prod_{j \in J}^{\times} \hat{f}_j(\hat{x}_{S_j})$ are a Fourier transform pair[4]. It then follows that for a given CFG (resp. MFG) representing $\prod_{j \in J}^{*} f_j(x_{S_j})$ (resp. $\prod_{j \in J}^{\times} f_j(x_{S_j})$), by letting the variable vertex indexed by $i$ represent $\hat{x}_i$, function vertex indexed by $j$ represent $\hat{f}_j$, and the product operation be defined as the "dual" operation $\times$ (resp. $*$), we arrive at a MFG (resp. CFG) representing $\prod_{j \in J}^{\times} \hat{f}_j(\hat{x}_{S_j})$ (resp. $\prod_{j \in J}^{*} \hat{f}_j(\hat{x}_{S_j})$) up to scale. We call the original and the resulting factor graph a pair of *dual factor graphs*.

---

[3]The notation $\prod^{\times}$ is then simply taken as $\prod$.

Another important property of the Fourier transform is the *slice-projection* (or *evaluation-marginalization*) duality. That is, for a given Fourier transform pair $f(x_1, x_2)$ and $\hat{f}(\hat{x}_1, \hat{x}_2)$ and any fixed configuration $\overline{x}_1$ of $x_1$, $f(\overline{x}_1, x_2)$ and $\sum_{\hat{x}_1} \hat{f}(\hat{x}_1, \hat{x}_2) \langle \hat{x}_1, \overline{x}_1 \rangle$ are a Fourier transform pair, where depending on the notions of the Fourier transform, the term $\langle \hat{x}_1, \overline{x}_1 \rangle$ may have various forms. For example, for the Fourier transform defined for functions on reals, $\langle \hat{x}_1, \overline{x}_1 \rangle$ is $e^{j 2\pi \hat{x}_1 \overline{x}_1}$, and for the Fourier transform defined for functions on $\{0, 1, 2, \ldots, N-1\}$, $\langle \hat{x}_1, \overline{x}_1 \rangle$ is $e^{j 2\pi \hat{x}_1 \overline{x}_1 / N}$, etc.

## 3 CFGS AS PROBABILITY MODELS

### 3.1 Modelling

It is well know that the summation of two independent random variables results in a convolution of their probability functions. That is, if independent random variables $X$ and $Y$ have probability functions $p_X(x)$ and $p_Y(y)$ respectively and random variable $Z := X + Y$, then the probability function $p_Z(z)$ of $Z$ is $p_X(z) * p_Y(z)$. This result generalizes to multiple random variables, as shown in Lemma 1.

**Lemma 1** *Let $(X, U)$ and $(V, Z)$ be two pairs of random variables and $(X, U) \perp\!\!\!\perp (V, Z)$. Suppose random variable $Y$ is defined as $Y := U + V$. Then the joint probability function $p_{XYZ}(x, y, z)$ of $(X, Y, Z)$ is*

$$p_{XYZ}(x, y, z) = p_{XU}(x, y) * p_{VZ}(y, z) \qquad (1)$$

*and the conditional probability function $p_{Y|XZ}(y, x, z)$ of $Y$ conditioned on $(X, Z)$ is*

$$p_{Y|XZ}(y, x, z) = p_{U|X}(y, x) * p_{V|Z}(y, z). \qquad (2)$$

---

[4]Depending on the notion of the Fourier transform, a scaling factor may be necessary.

*Proof:*

$$\begin{aligned}
p_{XYZ}(x,y,z) &= \sum_{u,v} p_{XYZUV}(x,y,z,u,v) \\
&= \sum_{u,v} p_{XUVZ}(x,u,v,z) p_{Y|UV}(y,u,v) \\
&= \sum_{(u,v):y=u+v} p_{XU}(x,u) p_{VZ}(v,z) \\
&= \sum_u p_{XU}(x,u) p_{VZ}(y-u,z) \\
&= p_{XU}(x,y) * p_{VZ}(y,z)
\end{aligned}$$

This proves (1). Noticing $X \perp\!\!\!\perp Z$, (2) can be proved similarly. □

In this lemma, random variables $U$ and $V$ are understood as *latent* or *hidden* random variables, which contribute to random variable $Y$ via summation. The moral of this lemma is "hidden sum implies convolution". Next lemma shows that the converse is also true, i.e., "convolution implies hidden sum".

We use the following short-hand notion in next lemma: for an arbitrary function $f(x,y)$, $f(x,+)$ refers to $\sum_y f(x,y)$.

**Lemma 2** *Let $X, Y$ and $Z$ be three random variables with joint probability function $p_{XYZ}(x,y,z) = f(x,y) * g(y,z)$ for some non-negative functions $f(x,y)$ and $g(y,z)$. Then*

1. $p_X(x) = \frac{f(x,+)}{\sum_x f(x,+)}$ *and* $p_Z(z) = \frac{g(+,z)}{\sum_z g(+,z)}$;

2. $X \perp\!\!\!\perp Z$; *and*

3. *there exist latent random variables $U$ and $V$ such that $(X,U) \perp\!\!\!\perp (Z,V)$ and $Y = U+V$, where*

$$p_{XU}(x,u) = \frac{f(x,u)}{\sum_{x,u} f(x,u)}$$

*and*

$$p_{VZ}(v,z) = \frac{g(v,z)}{\sum_{v,z} g(v,z)}.$$

*Proof:*

$$\begin{aligned}
p_X(x) &= \sum_{y,z} p_{XYZ}(x,y,z) \\
&= \sum_{y,z,u} f(x,y-u) g(u,z) \\
&= \sum_{z,u} f(x,+) g(u,z) \\
&= f(x,+) \sum_{u,z} g(u,z).
\end{aligned}$$

Since $p_X(x)$ is a probability function, $\sum_x p_X(x)$ must be 1. Then $\sum_{u,z} g(u,z)$ is necessarily $1/\sum_x f(x,+)$. Using the same argument for $p_Z(z)$, claim 1 is proved.

$$\begin{aligned}
p_{XZ}(x,z) &= \sum_y p_{XYZ}(x,y,z) \\
&= \sum_{y,u} f(x,y-u) g(u,z) \\
&= \sum_u f(x,+) g(u,z) \\
&= f(x,+) g(+,z) \\
&= \frac{f(x,+)}{\sum_x f(x,+)} \left( g(+,z) \sum_x f(x,+) \right) \\
&= p_X(x) p_Z(z).
\end{aligned}$$

The last equality above is by claim 1 and the fact $\sum_{u,z} g(u,z) = 1/\sum_x f(x,+)$ stated in proving claim 1. This proves claim 2.

Now we construct random variables $U$ and $V$ such that $(X,U) \perp\!\!\!\perp (V,Z)$, $p_{XU} = f(x,u)/\sum_{x,u} f(x,u)$ and $p_{VZ} = g(v,z)/\sum_{v,z} g(v,z)$. Construct random variable $Y' = U+V$. By Lemma 1, we have

$$\begin{aligned}
p_{XY'Z}(x,y',z) &= \frac{f(x,y') * g(y',z)}{\sum_{x,u} f(x,u) \sum_{v,z} g(v,z)} \\
&= f(x,y') * g(y',z)
\end{aligned}$$

That is, random variable $Y'$ is $Y$, and we have proved claim 3. □.

By the above lemmas, we have established a correspondence between summation of latent random variables and the convolutional factorization of the probability function of the observed variables. Now we proceed to give the general formalism of convolutional factor graphs as probabilistic models.

Let $U := \{1, 2, \ldots, |U|\}$ be a finite set indexing random variables $X_U$. Suppose that $U$ is partitioned into $m$ disjoint subsets $U(1), U(2), \ldots, U(m)$ such that $X_{U(i)} \perp\!\!\!\perp X_{U(j)}$ for every two distinct $i, j \in \{1, 2, \ldots, m\}$. Now re-partition the set $U$ into $K$ disjoint subsets, denoted by $V(1), V(2), \ldots, V(K)$ respectively, where the re-partitioning needs to satisfy the condition that if two distinct $\alpha$ and $\beta$ belong to the same $U(i)$, they can not simultaneously belong to the same $V(l)$ for any $l \leq K-1$. That is, if $\alpha, \beta \in V(l)$ for some $l \leq K-1$, then there is no $i \in \{1, 2, \ldots, m\}$ for which $\alpha, \beta \in U(i)$. Notice that this restriction does not apply to the elements $\alpha, \beta \in V(K)$. Indeed as will become apparent, the subset $V(K)$ indexes random variables in $X_U$ that are directly observed whereas every other subset $V(l), l \leq K-1$, indexes some random variables in $X_U$ that are latent and will contribute to the construction of another observed variable.

Now for each $l \leq K-1$, define random variable $X_{|U|+l} := \sum_{\alpha \in V(l)} X_\alpha$. This constructs a set of new random variables, each from a subset $X_{V(l)}$. For notational simplicity, define mapping $T : U \rightarrow \{1, 2, \ldots, |U| + K - 1\}$ by

$$T(\alpha) = \begin{cases} \alpha, & \alpha \in V(K) \\ |U| + l, & \alpha \in V(l), l \leq K-1. \end{cases}$$

Denote by $T(S)$ the image of any subset $S$ of $U$ under mapping $T$, then we have the following the theorem.

**Theorem 1** *The probability function of random variables $X_{T(U)}$ is $\prod_{i=1...m}^{*} p_{X_{U(i)}}(x_{T(U_i)})$.*

An extension of Lemma 1, the proof of this result is omitted. This theorem suggests that a convolutional factor graph can be constructed to represent the the joint probability function $p_{X_{T(U)}}(x_{T(U)})$ of observed random variables $X_{T(U)}$. It is helpful to first construct a so-called *chain graph* [7] to represent the underlying distribution of all random variables, including both observed and latent variables. Create an undirected graph with $m$ connectivity components where the $i^{th}$ connectivity component, $i = 1, \ldots, m$, forming a complete subgraph, includes vertices $X_{U(i)}$; for each $l \leq K-1$, create a new vertex $X_{|U|+l}$ and connect every vertex in set $X_{V(l)}$ to vertex $X_{|U|+l}$ with a directed edge (arrow pointing to $X_{|U|+l}$). These directed connections indicate specifically that $X_{|U|+l} = \sum_{\alpha \in V(l)} X_\alpha$. The resulting graph is then the chain graph representing the joint probability function $p_{X_{U \cup T(U)}}(x_{U \cup T(U)})$ of all variables. The CFG, representing the probability function of all observed variables only, consists of variable vertices $X_{T(U)}$ and function vertices are $p_{X_{U(i)}}$ for all $i = 1, 2, \ldots, m$, and each function vertex $p_{X_{U(i)}}$ connects to all variable vertices in $X_{T(U_i)}$. Specifically note that instead of taking some $x_\alpha$ as a variable, function $p_{X_{U(i)}}$ will take $x_{T(\alpha)}$ in place of $x_\alpha$ to form the convolutional product. We here give an example to illustrate this definition.

**Example 2** *Let $\{X_1, X_2\}, \{X_3, X_4, X_5, X_6\}$, and $\{X_7, X_8, X_9\}$ be three sets of random variables, where every two sets are independent. Let $X_{10} := X_2 + X_4 + X_8$ and $X_{11} := X_6 + X_9$. The underlying joint probability function $p_{X_1 \ldots X_{11}}(x_1, x_2, \ldots, x_{11})$ of all variables is represented by the chain graph in Figure 2 (left). Treating $X_2, X_4, X_6, X_8, X_9$ as latent variables, the joint probability function of $X_1, X_3, X_5, X_7, X_{10}, X_{11}$*

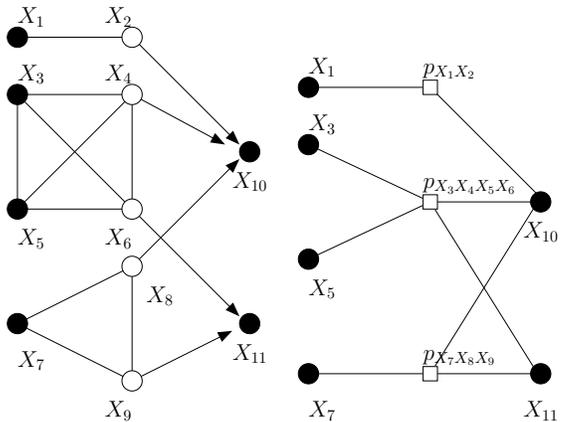

Figure 2: The chain graph (left) and CFG (right) of Example 2. The empty circles denotes the latent variables.

*equals $p_{X_1 X_2}(x_1, x_{10}) * p_{X_3 X_4 X_5 X_6}(x_3, x_{10}, x_5, x_{11}) * p_{X_7 X_8 X_9}(x_7, x_{10}, x_{11})$ according to Theorem 1. This gives rise to the CFG representation of $p_{X_1 X_3 X_5 X_7 X_{10} X_{11}}$ as of Figure 2(right).*

### 3.2 CFGs of Gaussian Densities

Consider jointly Gaussian random variables $(Y_1, Y_2, \ldots, Y_m)$ with arbitrary mean vector and covariance matrix $\mathbf{C}$. As follows, we will first create an undirected graph representing the structure of the covariance matrix $\mathbf{C}$. Let the graph consist of $m$ vertices, each representing a random variable $Y_i, i = 1, 2, \ldots, m$; vertex $Y_i$ and $Y_j$ are connected by an edge if $\mathbf{C}_{ij} \neq 0$. This graph is also known as the *covariance graph* (see, e.g., [8] and [9]) of the Gaussian density. Let $Q$ denote the set of maximal cliques of the covariance graph. Then the following theorem can be proved.

**Theorem 2** *The Gaussian density of $(Y_1, Y_2, \ldots, Y_m)$ factors convolutionally as $\prod_{C \in Q}^{*} f_C(y_C)$, where each $f_C$ is a $|C|$-dimensional Gaussian density.*

This result suggests that every multi-variate Gaussian density has a convolutional factorization, and further that the factorization structure corresponds precisely to the covariance matrix. It then follows that the CFG representation of a Gaussian density has a structural correspondence with the covariance matrix. Due to length constraint, we omit the proof of this result and simply present an example.

**Example 3** *Let $\mathbf{C}$ be the covariance matrix of jointly Gaussian random variables $(Y_1, Y_2, \ldots, Y_6)$. Suppose*

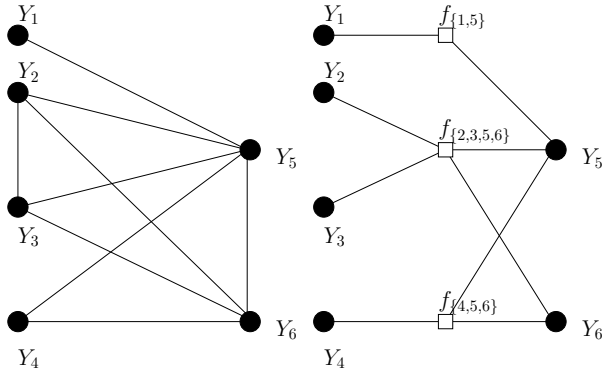

Figure 3: The covariance graph (left) and the CFG (right) of the Gaussian density in Example 3.

that $\mathbf{C}$ has the following structure

$$\mathbf{C} = \begin{bmatrix} \times & 0 & 0 & 0 & \times & 0 \\ 0 & \times & \times & 0 & \times & \times \\ 0 & \times & \times & 0 & \times & \times \\ 0 & 0 & 0 & \times & \times & \times \\ \times & \times & \times & \times & \times & \times \\ 0 & \times & \times & \times & \times & \times \end{bmatrix},$$

where $\times$ denotes a non-zero element. Figure 3 (left) shows the covariance graph, and Figure 3 (right) shows the CFG representing the Gaussian density.

### 3.3 CFGs of IF Models

Summation of independent latent random variables appear in a large class of probability models, beyond those for Gaussian densities. We stress that in the setup for CFG modelling, there is no restriction on the functional forms of the involved probability functions, while the only restriction is that the independent latent variables contribute to forming the observed variables via summation, or equivalently, via a linear transformation. Here we give another example, the CFG representation of the *independent factor* (IF) generative model ( [10] and [11]).

The IF model consists of a set of hidden *source* variables $X_1, X_2, \ldots, X_m$, mutually independent, a set of *sensor* variables $Y_1, Y_2, \ldots, Y_L$, and an additive noise vector $(U_1, U_2, \ldots, U_L)$ independent of the sources; the observed sensor vector is modelled generatively as

$$(Y_1, \ldots, Y_L)^T = \mathbf{H} \, (X_1, \ldots, X_m)^T + (U_1, \ldots, U_L)^T,$$

where $\mathbf{H}$ is an $L \times m$ matrix. Typically, noise $(U_1, \ldots, U_L)$ is specified as jointly Gaussian with zero mean. A judicious choice of the density $p_{X_i}$ of each source variable $X_i$, as shown in [11], is a mixture-of-Gaussian density. Clearly, such a model generalizes the models of *factor analysis*, *principal component analysis* and *independent component analysis*. The BN view of the IF model is shown in Figure 4 (top).

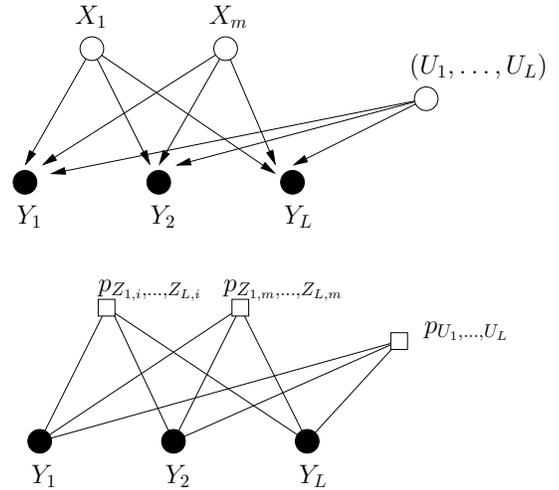

Figure 4: The BN (top) and CFG (bottom) representation of an IF model.

To construct the CFG representation of the IF model, for each source variable $X_i$, we need to first introduce a set of variables $Z_{j,i} := \mathbf{H}_{j,i} X_i, j = 1, \ldots, L$, to represent the contribution of the $i^{\text{th}}$ source to the $j^{\text{th}}$ sensor. This defines a joint probability function $p_{Z_{1,i} \ldots Z_{L,i}}(z_{1,i}, \ldots, z_{L,i})$ of $(Z_{1,i}, \ldots, Z_{L,i})$. That is, $p_{Z_{1,i} \ldots Z_{L,i}}(z_{1,i}, \ldots, z_{L,i})$ evaluates to $p_{X_i}(x_i)$ at $(z_{1,i}, \ldots, z_{L,i}) = (\mathbf{H}_{1,i} x_i, \ldots, \mathbf{H}_{L,i} x_i)$, and evaluates to 0, otherwise. Then by Theorem 1, the joint probability function $p_{Y_1, \ldots, Y_L}(y_1, \ldots, y_L)$ of the sensor variables is $p_{U_1, \ldots, U_L}(y_1, \ldots, y_L) *$ $\prod_{i=1 \ldots m}^{*} p_{Z_{1,i}, \ldots, Z_{L,i}}(y_1, \ldots, y_L)$, and can be represented by the CFG in Figure 4 (bottom).

### 3.4 Marginal Independence on CFG Models

**Proposition 1** *On a CFG representation of the joint probability function $X_V$, if disjoint variable vertex sets $X_A, X_B$ and $X_S$ satisfy that $X_S$ separates $X_A$ from $X_B$, then $X_A \perp\!\!\!\perp X_B$.*

This result is a straight-forward extension of Lemma 2 and we also omit its proof. Clearly, this property of CFGs resembles, in a "dual" way, the global Markov property of a MFG or MRF. That is, on a MFG, *conditioning* on a cut-set induces independence, whereas on a CFG, *marginalization* over a cut-set induces independence. As will become more evident in next section, marginalization and conditioning are in fact duals under the Fourier transform.

## 4 INFERENCE

Given the probability function $p_{X_V}(x_V)$ of a set of random variables $X_V$, an inference problem may be regarded as various versions of the following generic problem: for certain disjoint subsets $M$, $E$ and $R$ of the index set $V$ where $M \cup E \cup R = V$ and for a given configuration $\overline{x}_E$ of variables $x_E$, determine

$$\sum_{x_M} p_{X_V}(x_M, x_R, \overline{x}_E).$$

That is, an inference problem typically involves either the *evaluation* of a probability function or the *marginalization* of a probability function, or both. Notice that marginalization is a necessary procedure of computing any *marginal distribution*, whereas evaluation is a necessary procedure of computing any *conditional distribution*.

Since the popular choices of graphical models — BNs and MRFs — can both be converted to a MFG, we will first review the solution of the generic inference problem on a MFG. We then discuss solving the generic inference problem on a CFG, which exhibits a clear algorithmic symmetry with the MFG case.

### 4.1 Inference on MFGs

Let the MFG represent $p_{X_V}(x_V) = \prod_{j=1...m} f_j(x_{S_j})$. Then evaluating $p_{X_V}(x_V)$ at $x_E = \overline{x}_E$ gives rise to the function

$$p_{X_V}(x_M, x_R, \overline{x}_E) = \prod_{j=1...m} f_j(x_{S_j \setminus E}, \overline{x}_{S_j \cap E}),$$

where $\overline{x}_{S_j \cap E}$ is the component of $\overline{x}_E$ indexed by $S_j \cap E$. This essentially defines a new product function, where the factors are $f_j(x_{S_j \setminus E}, \overline{x}_{S_j \cap E}), j = 1, \ldots, m$. We may then represent this product by a different MFG with reduced structure, and the generic inference problem becomes a pure marginalization problem. That is, we may without loss of generality consider the generic inference problem on a MFG as computing $\sum_{x_M} F_{X_V}(x_V)$, for some function $F_{X_V}(x_V) = \prod_{j=1...m} f_j(x_{S_j})$.

This problem can be in fact solved by an algorithm on the MFG, equivalent to the Elimination algorithm [12] on a BN. We now describe this algorithm, which we refer to as the MFG-Elimination algorithm. Select an ordering of the elements in set $M$, and identify each element of $M$ by its order $1, 2, \ldots, |M|$. We will not be concerned with the optimal ordering of the elements of $M$, it is however preferable to let the indices of the leaf vertices to appear first in the ordering. We will use $\mathcal{N}(\cdot)$ to denote the set of adjacent vertices of any given vertex. The MFG-Elimination algorithm is then an algorithm that successively reduces the MFG, given as follows.

```
for i = 1 to |M|
{
f_{m+i}(N(x_i) \ {x_i}) := ∑_{x_i} ∏_{j:f_j ∈ N(x_i)} f_j(x_{S_j});
remove {f_j : f_j ∈ N(x_i)} and x_i from MFG;
add f_{m+i} to MFG;
}
```

At this end, the resulting MFG represents the desired function $\sum_{x_M} F_{X_V}(x_V)$. We remark that in the MFG-Elimination algorithm, we in fact do not require that $F_{X_V}$ be a probability function or an evaluated probability function, i.e., for an arbitrary function $F_{X_V}$ of variables $x_V$, we can use the MFG-Elimination algorithm to solve for $\sum_{x_M} F_{X_V}(x_V)$.

### 4.2 Inference on CFGs

Let the CFG represent $p_{X_V}(x_V) = \prod_{j=1...m}^{*} f_j(x_{S_j})$. Then it can be shown, in a way similar to the proving claim 2 of Lemma 2, that marginalizing $p_{X_V}(x_V)$ over $x_M$ gives rise to the function

$$\sum_{x_M} p_{X_V}(x_M, x_R, x_E) = \prod_{j=1...m}^{*} \sum_{x_{M \cap S_j}} f_j(x_{S_j}).$$

This essentially defines a new convolutional product function, where the factors are $\sum_{x_{M \cap S_j}} f_j(x_{S_j}), j = 1, \ldots m$. We may then represent this product by a different CFG with reduced structure, and the generic inference problem becomes a pure evaluation problem. That is, we may without loss of generality consider the generic inference problem on a CFG as computing $F_{X_V}(x_{V \setminus E}, \overline{x}_E)$, for some function $F_{X_V}(x_V) = \prod_{j=1...m}^{*} f_j(x_{S_j})$.

Similar to marginalization on a MFG, evaluation problem on a CFG can be solved by successively reducing structure of the CFG. In what follows we describe what we call the CFG-Elimination algorithm for this purpose. Select an ordering of the elements in set $E$, and identify each element of $E$ by its order $1, 2, \ldots, |E|$. Again, we prefer an ordering such that the indices of the leaf vertices appear first in the ordering. The CFG-Elimination algorithm is then given as follows.

```
for i = 1 to |E|
{
```
$$f_{m+i}(\mathcal{N}(x_i) \setminus \{x_i\}) := \left[\prod_{j:f_j \in \mathcal{N}(x_i)}^{*} f_j(x_{S_j})\right]_{x_i = \overline{x}_i};$$
```
remove {f_j : f_j ∈ N(x_i)} and x_i from CFG;
add f_{m+i} to CFG;
}
```

At this end, the resulting CFG represents the desired function $F_{X_V}(x_{V \setminus E}, \overline{x}_E)$.

In fact one can show that this algorithm is identical to representing the underlying distribution including the latent variables by a MFG and following a certain order to perform evaluation at a subset of observed vertices and marginalization over all latent vertices. Indeed, the CFG-Elimination algorithm is not the most efficient way of computing $F_{X_V}(x_{V \setminus E}, \overline{x}_E)$, particularly if each variable takes values from a large set of configurations. Our description of CFG-Elimination is mainly to demonstrate the algorithmic duality between marginalization problems and evaluation problem. Exploiting the evaluation-marginalization duality and the convolution theorem of the Fourier transform, a more efficient way for computing $F_{X_V}(x_{V \setminus E}, \overline{x}_E)$ can be performed via the Fast Fourier Transform (FFT). More specifically, note that the Fourier transform of $F_{X_V}(x_{V \setminus E}, \overline{x}_E)$ is

$$\sum_{\hat{x}_E} \hat{F}(\hat{x}_V) \langle \hat{x}_E, \overline{x}_E \rangle = \sum_{\hat{x}_E} \prod_{j=1...m} \hat{f}_j(x_{S_j}) \langle \hat{x}_{S_j \cap E}, \overline{x}_{S_j \cap E} \rangle.$$

That is, we can take the FFT of each convolutional factor $f_j$ [5], define $g_j(\hat{x}_{S_j}) := \hat{f}_j(x_{S_j}) \langle \hat{x}_{S_j \cap E}, \overline{x}_{S_j \cap E} \rangle$, construct an MFG representing $\prod_{j=1...m} g_j(\hat{x}_{S_j})$ and compute $\sum_{\hat{x}_E} \prod_{j=1...m} g_j(\hat{x}_{S_j})$ on the MFG by the MFG-Elimination algorithm. By taking the inverse FFT of the resulting function we obtain the desired function $F(x_{V \setminus E}, \overline{x}_E)$. We note that the MFG representing $\prod_{j=1...m} g_j(\hat{x}_{S_j})$ has precisely the same structure as the CFG representing $F(x_V)$. As is typically expected with FFT based implementations, it can be shown that using the MFG-Elimination algorithm via the FFT, the computational saving for each computation of convolution [6] is by a factor of approximately $A/\log A$, where $A$ is the number of values that a variable can take.

We remark that it is also possible to develop algorithms for CFGs that are analogous to the Sum-Product or Junction Tree algorithm on MFGs. Here we omit this discussion for simplicity.

## 5 CONCLUDING REMARKS

Building upon a previous work, this paper introduces CFGs as a new class of probabilistic models. We show that CFGs are natural representations of probability functions of observed random variables which are obtained as a linear transformation of a set of independent latent random variables. Explicitly identifying the convolutional factorization of the modelled probability functions, CFG models are facilitated with the tools of Fourier analysis, which may provide convenience for either analysis or computation.

---

[5] More precisely, it is only necessary to take the "partial" FFT for each $f_j(x_{S_j})$ with respect to the non-leaf variables. We here choose not to elaborate on this subtlety.

[6] When using the CFG-Elimination algorithm (or the equivalent algorithms without using the CFG representation), convolution arises explicitly or implicitly when eliminating each non-leaf variable vertex and when computing the function represented by the final resulting CFG.